# On-the-Fly Learning in a Perpetual Learning Machine

Andrew J.R. Simpson [#1]

[#] *Centre for Vision, Speech and Signal Processing, University of Surrey, UK*
[1] `Andrew.Simpson@Surrey.ac.uk`

*Abstract*—**Despite the promise of brain-inspired machine learning, deep neural networks (DNN) have frustratingly failed to bridge the deceptively large gap between** *learning* **and** *memory*. **Here, we introduce a** *Perpetual Learning Machine*; **a new type of DNN that is capable of brain-like dynamic 'on the fly' learning because it exists in a** *self-supervised* **state of** *Perpetual Stochastic Gradient Descent*. **Thus, we provide the means to unify** *learning* **and** *memory* **within a machine learning framework. We also explore the elegant duality of** *abstraction* **and** *synthesis*: **the** *Yin* **and** *Yang* **of deep learning.**

## I. INTRODUCTION

It is an embarassing fact that while deep neural networks (DNN) are frequently compared to the brain, and even their performance found to be similar in specific static tasks, there remains a critical difference; DNN do not exhibit the fluid and dynamic learning of the brain but are static once trained. For example, to add a new class of data to a trained DNN it is necessary to add the respective new training data to the pre-existing training data and re-train (probably from scratch) to account for the new class. By contrast, learning is essentially additive in the brain – if we want to learn a new thing, we do.

Thus, whilst there is little doubt that the *learning* of the brain and machine *learning* are essentially the same, the learning of the brain involves the emergent phenomenon of *memory* which has failed to emerge from machine learning. Indeed, recent machine-inspired approaches to 'memory' have involved explicit add-on *storage* facilities [e.g., 1] which explicitly discriminate between *learning* (training – i.e., of weights) and *memory* (storage – i.e., of data). Thus, the problem has been brushed under the carpet.

In this article, we describe a novel form of supervised learning model, which we call a *Perpetual Learning Machine*, which gives rise to the basic properties of *memory*. Our model involves two DNNs, one for *storage* and the other for *recall*. The *storage* DNN learns the classes of some training images. The *recall* DNN learns to synthesise the same images from the same classes. Together, the two networks hold, encoded, the training set. We then place these pair of DNNs in a self-supervised and homeostatic state of *Perpetual Stochastic Gradient Descent* (PSGD). During each step of PSGD, a random class is chosen and an image synthesised from the *recall* DNN. This randomly synthesised image is then used in combination with the random class to train both DNNs via non-batch SGD. I.e., the PSGD is driven by training data that is *synthesised* from *memory* according to random classes. We next demonstrate that new classes may be learned on the fly by introducing them, via 'new experience' SGD steps, into the path of PSGD. Over time, new classes are assimilated without disruption of earlier learning and hence we demonstrate a machine which both *learns* and *remembers*.

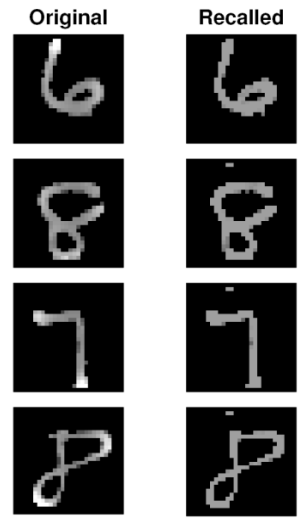

**Fig. 1. Recall of training images.** On the left are plotted MNIST digits and on the right are plotted the same digits synthesised using the *recall* DNN.

## II. METHOD

We chose the well-known MNIST hand-written digit dataset [2]. First, we unpacked the images of 28x28 pixels into vectors of length 784. Example digits are given in Fig. 1. Pixel intensities were normalized to zero mean.

*Perpetual Memory.* In order to test the idea of perpetual memory, through perpetual learning, we required our model to learn to identify a collection of images. We took the first 75 of the MNIST digits and assigned each to an arbitrary class (this is arbitrary associative learning). This gave 75 unique classes, each associated with a single, specific digit. The task of the model was to recognise the images and assign to them the correct (arbitrary) classes. We split the 75 digits into 50 'learn during training' examples and 25 'learn later on the fly' test examples. The first 50 training examples were learned with typical SGD and discarded. *Hence, they were not available for later use during assimilation of additional classes.* The latter 25 examples were held back for insertion during PSGD.

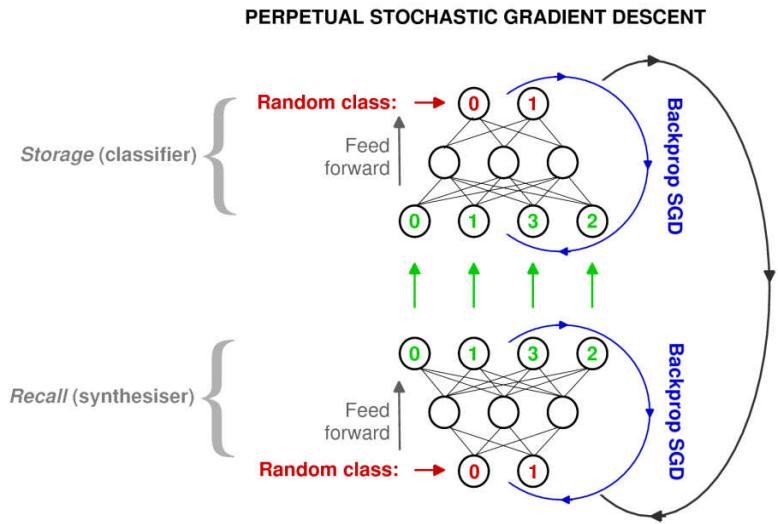

**Fig. 2. Self-supervision: PSGD schematic diagram.** For each iteration of PSGD, a random class is chosen and from this input the *recall* DNN is used to synthesise the respective training image (from memory). This recalled training image is then used with the random class to train *both* networks for a single step of backprop SGD.

*Storage and Recall.* We instantiated two DNN; the *storage* DNN was a typical classifier of size 784x100x75, with the softmax output layer corresponding to the 75-way classification problem. In addition to the 50 training classes, the 75 possible classes provides for 25 redundant (unused) classes to be learned later. The *storage* DNN took images as input and produced classes as output. The *recall* DNN was of size 75x100x784, took classes as input and synthesised the training images at output. Both DNNs used biased sigmoids [3] throughout (with zero bias in the output layer).

The *storage* and *recall* DNNs were trained, independently, using only the first 50 images for 100 full-sweep iterations of typical non-batch SGD [4,5]. Training was performed (regularised) using parallel (100x) dither w/ dropout [as in 5]. After 100 iterations, classification error was converged at 0.02% (1 mistake in 50) for the *storage* DNN, and at 0.02% for the *storage* DNN fed with the output of the *recall* DNN charged with synthesising the images of the respective test classes. Hence, the recall was suitably robust and was more or less visually indistinguishable from the original training images. Fig. 1 plots some example digits recalled (synthesised) using the *recall* DNN.

*Perpetual Stochastic Gradient Descent.* Once the pair of DNN were independently trained on the 50 training images, the training images were discarded and the DNN were placed in a mutually recurrent and perpetual circuit. Fig. 3 shows a schematic diagram of PSGD; For each iteration of the PSGD, a random class was chosen (from the total 75 possible). Next, using this random class, a respective image was synthesised using the *recall* DNN. This synthetic image was then combined with the random class and used together to train both DNNs in parallel (via non-batch SGD [5]). I.e., given the random seed, the *recall* DNN synthesised – from *memory* – the relevant training image and used it for *self-supervision*. This step of non-batch SGD also employed parallel dither w/dropout (100x). As in [4,5], all dither was random noise of zero mean and unit scale and dropout [6,5] was 50%.

*Introduction and learning of new classes.* After 1000 iterations of PSGD, during which time homeostatic convergence had occurred, we enter a 2000-iteration on-the-fly-learning epoch. In this epoch, new classes (of the 25) were introduced at random. At each iteration of PSGD, one of the 25 new digits was chosen (at random) and both DNN were trained for an additional single step (using 100x parallel dither w/ dropout) using the respective training data. Thus, the new training data were introduced into the path of the PSGD.

Finally, at the end of the 2000-iterations on-the-fly-learning epoch, the introduction of new classes was stopped and the PSGD continued for a further 2000 iterations, during which time homeostatic convergence was reinstated (i.e., emerged).

To test the efficacy, on-the-fly dynamics and robustness of this perpetual memory, the classification accuracy of the *storage* DNN was tested at each iteration of PSGD. To do this, first the test classes were passed through the *recall* DNN to synthesise the 'remembered' images. These images were then classified using the *storage* DNN and the accuracy computed. At each PSGD iteration, to test retention of the pre-learned training set, test error was computed for the 50 training examples. In addition, to test the assimilation of the new classes, test error was computed on the 25 new classes. Finally, test error was computed over the whole set. This gave three dynamic measures of memory and recall that could be plotted as a function of time.

III. RESULTS

Fig. 3a plots the recall accuracy (classification error rate) of the *storage* DNN (via the *recall* DNN, as described above) as a function of PSGD iterations. Fig. 3b plots the recall accuracy of the *storage* DNN via the *recall* DNN (as described above) as a function of PSGD iterations. Separate error functions are plotted for the training set of 50 (blue), the on-the-fly test set of 25 (green) and the entire set of 75 (red). In the case of the training set error function (blue), the first 1000 iterations demonstrate the homeostasis of perpetual learning – *memory*. The first 1000 iterations also demonstrate that classification accuracy in the on-the-fly test set (green) is at zero (error rate of 1) for the *storage* DNN (Fig. 3a) and is around chance level for the *storage*-via-*recall* DNN (Fig. 3b) and the combined full-set error rates are respective weighted averages of the two (red).

At 1000 iterations begins the 2000-iteration on-the-fly-learning epoch, where new classes are learned. During this epoch, each step of PSGD additionally involves a SGD step taken with a randomly selected element of the *new* test set (25) concurrently with the continuing self-supervised PSGD (random steps of self-supervision). Initially, the training set (blue) and full set (red) error functions deflect upwards as the impact of the new classes causes adjustment of the paths of the pre-established training set. At the same time, classification accuracy on the on-the-fly test set (green) begins to fall rapidly until it converges. Within a few hundred iterations of PSGD homeostasis is reinstated, as is reflected across the various error functions. At the 3000 iterations point (the end of the on-the-fly-learning epoch) the introduction of new data ceases with no obvious effect (even for the new test set error), illustrating a homeostasis which results from all 75 classes being fully assimilated into the perpetual *memory*. Essentially, 25 new image classes were learned on the fly, and retained after the on-the-fly-learning epoch, with no significant loss in accuracy for the original, pre-learned 50-strong training set.

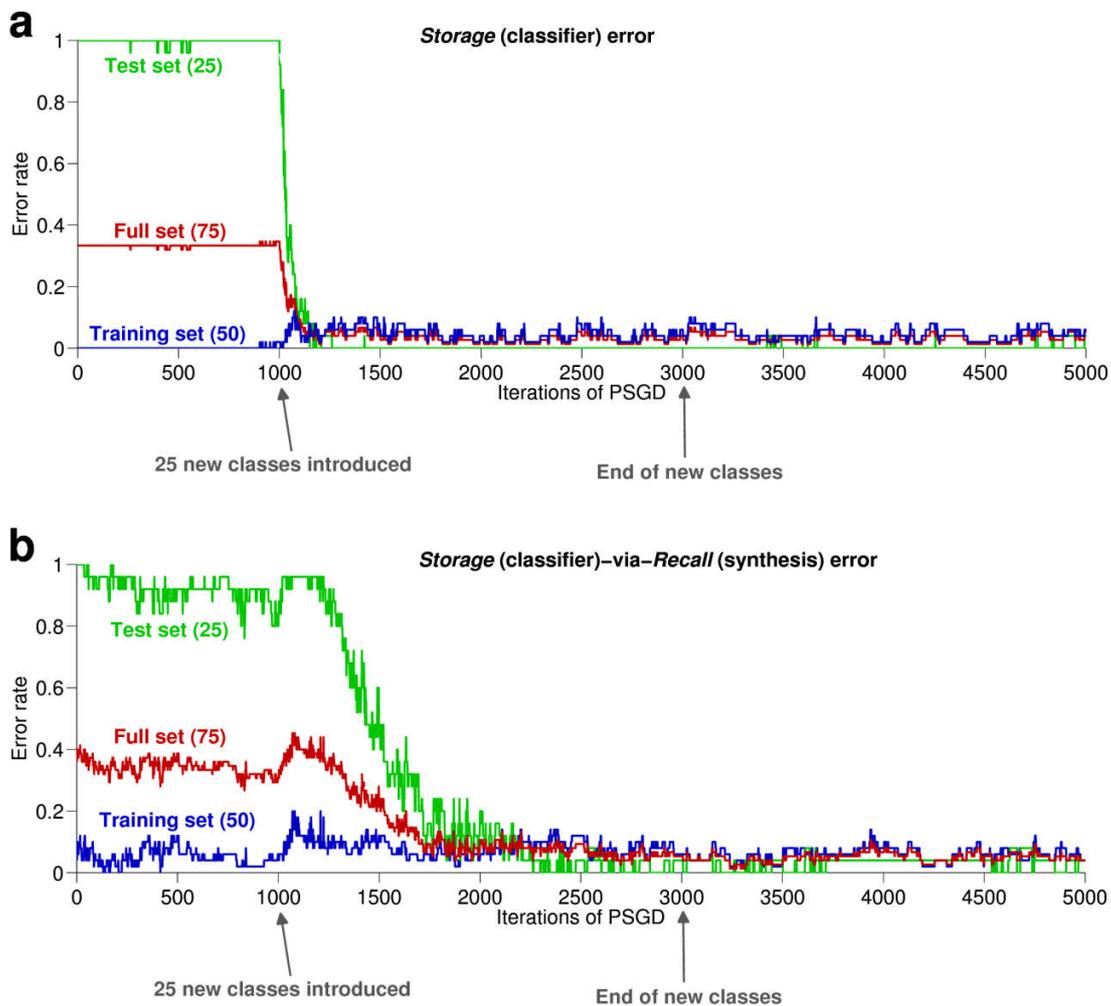

**Fig. 3. Perpetual memory and on-the-fly learning via PSGD.** Various *recall* DNN error functions of PSGD iterations. **a** plots error for the *storage* DNN and **b** plots error for the *storage* DNN fed with images recalled from the *recall* DNN. The machine is in homeostasis for the first 1000 PSGD iterations, then begins the 2000-iteration on-the-fly-learning epoch within which randomly selected elements from the new test set are learned during the concurrent PSGD. Finally, at the 3000-iteration point, the on-the-fly-learning phase ends and homeostasis is reinstated.

We note that (data not shown) no aspect of the current model trains successfully without regularisation via parallel dither (w/ dropout) [5]. The attempt to train on the training set without parallel dither resulted in sufficiently poor performance that any form of PSGD did not succeed at all. Even if the initial training was done with parallel dither, if the PSGD was undertaken without parallel dither (i.e., either without any regularisation or with dropout) the result was a rapid explosion of error that did not recover. Hence, it appears that the regularisation provided by parallel dither (w/ dropout) is critical.

## IV. DISCUSSION AND CONCLUSION

We have demonstrated a powerful new class of self-supervised deep neural network – a *Perpetual Learning Machine* -, capable of both maintaining perpetual memory and assimilating new classes into perpetual memory without the use of the original training set. To achieve this we have placed two feed-forward DNNs – one for *storage* (classification) and the other for *recall* (synthesis) - in a recurrent and self-supervised state of *Perpetual Stochastic Gradient Descent*. Each iteration of PSGD involves parallel backpropagated SGD driven by random recall of the training set. We have demonstrated that new classes can be assimilated into the perpetual memory by inserting concurrent SGD steps training for new image classes. I.e., we have demonstrated both on-the-fly learning and the persistence of what is learned on-the-fly.

In summary, we have shown how *memory* emerges from *learning* and hence we have provided a means to unify learning and memory within a machine learning framework. Unlike conventional 'memory' (storage) devices attached to neural networks (e.g., LSTM [1]), this new class of Perpetual Learning Machine is able to store, retain, recall and add *memories* in perpetuity. Furthermore, unlike the conventional storage-based 'memory' devices (e.g., [1]), the present unified architecture represents both learning and memory through the states of its weights. Hence, it seems possible that a similar principle may be responsible for both learning and memory within the brain.

## V. APPENDIX: THE YIN AND YANG OF DEEP LEARNING

At the heart of deep learning is the question: *What is the inverse (or opposite) of abstraction*? The answer is as simple as it is elegant: *synthesis*.

Deep neural networks are typically interpreted in one of two ways. The first is the Universal Approximation Theorem interpretation [7-11]. The second is the Probabilistic interpretation [12-15] and usually involves inference. The former interpretation [7-11] is concerned with characterising the nature of the function and is agnostic to the data or what the function might be useful for. The latter interpretation [12-15] is concerned with the utility of the function and is agnostic to the processing of the data by the function.

More recently, the discrete signal processing (DSP) interpretation [3-5,16-18] has been introduced. Critically, in contrast to the two prevailing interpretations (Universal Approximation and Probabilistic Inference), the DSP interpretation is concerned with the data (i.e., the signal) and *how* it is processed by the function. This interpretation is therefore agnostic to the function itself and to the utility of the function. Thus, we may leverage pure physics to interrogate both the data and the process of deep learning.

***The Discrete Signal Processing Interpretation***. In the DSP interpretation of deep neural networks [3-5,16-18], the DNN is constructed of linear filters, followed by nonlinear activation functions. The linear filters are selective of *features*. The nonlinear activation functions act as demodulators of the *features* [3,16]. Demodulation occurs in the activation function as a result of *rectification* [3,16]. Demodulation is the process whereby we obtain the magnitude of the variance of the features of a signal [see 3,16]. In the opposite direction, we may restore the original signal by convolving the demodulated magnitude with the feature – this is *synthesis* [18,19].

We can sanity check this intuition with the following simple thought experiment. Let us consider the bounds on the inverse of abstraction – synthesis – by considering what happens if we try to synthesise a negative quantity of apples; In our physical reality, this does not make sense. This *Yin-Yang* intuition provides an interesting window into the nature of what is going on during learning (training) in a deep neural network.

Figure A4 gives a schematic diagram, depicting the duality of abstraction and synthesis as it flows in opposing directions for training (feed back) and testing (feed forward). First, let us consider the case of a classifier – a feed-forward DNN whose task is to learn to *abstract* classes from images (upper diagram, Fig. A4). When we apply back-propagation gradient descent we first compute the error at the output layer. This error exists at the abstract level of 'classes' and, so, in order to propagate the error backwards towards the original feature-space of the input data, we must implicitly *synthesise the weights*.

Now, let us consider the opposite case of a synthesiser (as in Fig. 2) – a feed-forward DNN whose task is to synthesise images from abstract classes (the lower diagram of Fig. A4). When we apply back-propagated gradient descent, we first compute the error in the non-abstract (i.e., native) feature space of the original image. Then, in order to propagate this error backwards (*towards* a higher level of abstraction of classes at the input layer), we are *abstracting* the *error* signal as we propagate downwards from the output layer. Thus, we apply demodulation, through the activation function, during back propagation in this network and synthesis during forward propagation.

These two opposing cases provide an elegant duality of intuition; In both networks there are opposing flows of abstraction and synthesis during feed-forward and feed-backward (back propagation). In both processes, abstraction occurs by demodulation and synthesis occurs by convolution (see [19]).

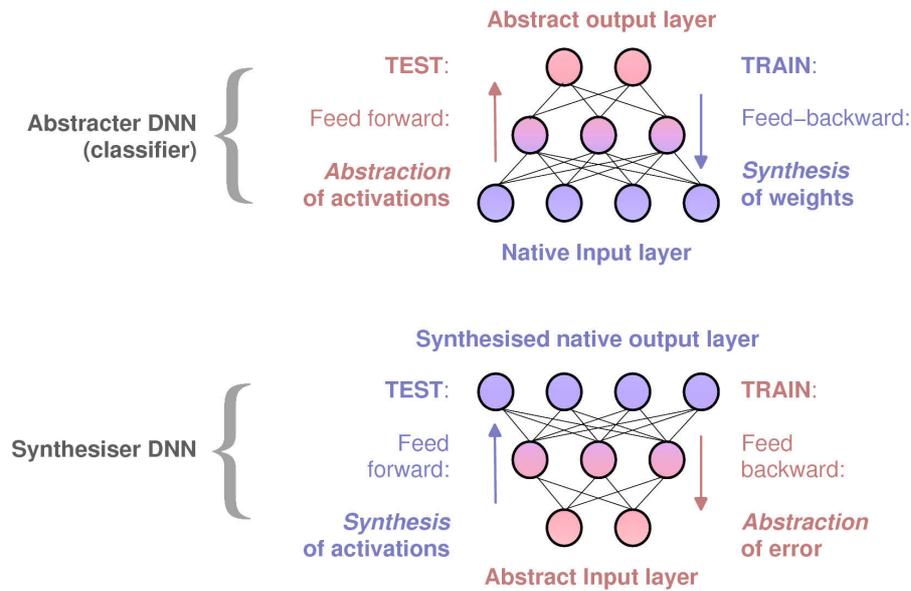

**Fig. A4. Yin and Yang: The duality of abstraction and synthesis.** Schematic diagram indicating the flow of abstraction and synthesis for a classifier (which performs abstraction in feed-forward mode) and a synthesiser (which performs synthesis in feed-forward mode). Both feature both abstraction and synthesis but in opposing directions. This directionality is determined by the data, not the architecture.

A further thought experiment might be useful to understand how the data place bounds on the depth of learning that can be attained with an arbitrary deep neural network. Let us consider what happens when we demodulate a pure magnitude:- because there is no variance to act as carrier signal (see [3,16]), when we demodulate there is nothing left for us to apply selective filters to. Thus, we cannot demodulate beyond the highest (most abstract) available carrier (see [19] for conservation of energy during demodulation).

*Insights*. This leads to two specific insights into the nature of deep learning:

*Insight 1*: It is the data which define the process as either synthesis or abstraction during both feed-forward and feed-backward directions of propagation.

*Insight 2*: The data define the depth we may *abstract to* or *synthesise from* and hence the data place bounds on the depth of learning that is possible (not the architecture – e.g., see Universal Approximation Theorem [7-11]).

Finally, given that essentially the same argument may be made for the signal processing of the brain, it may be that the *Yin* and *Yang* of deep learning lays the ground work for a redefinition of (artificial) intelligence itself, based on the physics of the known universe and the limits of our sensory apparatus. Indeed, if the brain is interpreted as a universal function approximation device, then intelligence may be limited not by the brain but by the degree of abstraction in the physical world which may be captured by our sensory apparatus.

ACKNOWLEDGMENT

AJRS did this work on the weekends and was supported by his wife and children.

REFERENCES

[1] Hochreiter S, Schmidhuber J (1997) "Long short-term memory", Neural computation 9: 1735-1780.
[2] LeCun Y, Bottou L, Bengio Y, Haffner P (1998) "Gradient-based learning applied to document recognition", Proc. IEEE 86: 2278–2324.
[3] Simpson AJR (2015) "Abstract Learning via Demodulation in a Deep Neural Network", arxiv.org abs/1502.04042.
[4] Simpson AJR (2015) "Dither is Better than Dropout for Regularising Deep Neural Networks", arxiv.org abs/1508.04826.
[5] Simpson AJR (2015) "Parallel Dither and Dropout for Regularising Deep Neural Networks", arxiv.org abs/1508.07130.
[6] Hinton GE, Srivastava N, Krizhevsky A, Sutskever I, Salakhutdinov R (2012) "Improving neural networks by preventing co-adaptation of feature detectors", The Computing Research Repository (CoRR), abs/1207.0580.
[7] Cybenko G (1989) "Approximations by superpositions of sigmoidal functions", Mathematics of Control, Signals, and Systems 2: 303-314.
[8] Csáji BC (2001) "Approximation with artificial neural networks", Faculty of Sciences, Etvs Lornd University, Hungary, 24.
[9] Hornik K (1991) "Approximation Capabilities of Multilayer Feedforward Networks", Neural Networks 4: 251–257.
[10] Haykin S (1998). "Neural Networks: A Comprehensive Foundation", Vol. 2, Prentice Hall.
[11] Hassoun M. (1995) "Fundamentals of Artificial Neural Networks", MIT Press.
[12] LeCun Y, Bottou L, Bengio Y, Haffner P (1998) "Gradient-based learning applied to document recognition", Proc. IEEE 86: 2278–2324.
[13] Hinton GE, Osindero S, Teh YW (2006) "A Fast Learning Algorithm for Deep Belief Nets", Neural Computation 18: 1527–1554.
[14] Bengio Y (2009) "Learning deep architectures for AI", Foundations and trends in Machine Learning 2: 1-127.
[15] LeCun Y, Bengio Y, Hinton GE (2015) "Deep learning", Nature 521: 436–444.
[16] Simpson AJR (2015) "Taming the ReLU with Parallel Dither in a Deep Neural Network", arxiv.org abs/1509.05173
[17] Simpson AJR (2015) "Over-Sampling in a Deep Neural Network", arxiv.org abs/1502.03648.
[18] Simpson AJR (2015) "Deep Transform: Error Correction via Probabilistic Re-Synthesis", arxiv.org abs/1502.04617.
[19] Bruna J, Mallat S (2013) "Invariant scattering convolution networks", Pattern Analysis and Machine Intelligence, IEEE Transactions on, 35: 1872-1886.